\documentclass[10pt,twocolumn,letterpaper]{article}

\usepackage[pagenumbers]{cvpr} %

\usepackage{color}
\usepackage{amsmath}
\usepackage{amssymb}
\usepackage{amsfonts}
\usepackage{multirow, varwidth}
\usepackage{epsfig}
\usepackage{verbatim}
\usepackage{dsfont}
\makeatletter
\newif\if@restonecol
\makeatother
\usepackage{xr}
\usepackage[amsmath,thmmarks]{ntheorem}
\usepackage{booktabs}
\usepackage{comment}
\usepackage{paralist}

\DeclareRobustCommand\onedot{\futurelet\@let@token\@onedot}
\def\onedot{.} %
\def\eg{\emph{e.g}\onedot} 
\def\ie{\emph{i.e}\onedot}

\def\etal{\emph{et al}\onedot}

\usepackage{graphicx}
\usepackage{amsmath}
\usepackage{amssymb}
\usepackage{booktabs}

\usepackage[pagebackref,breaklinks,colorlinks]{hyperref}

\usepackage[capitalize]{cleveref}
\crefname{section}{Sec.}{Secs.}
\Crefname{section}{Section}{Sections}
\Crefname{table}{Table}{Tables}
\crefname{table}{Tab.}{Tabs.}

\def\cvprPaperID{15} %
\def\confName{AICC}
\def\confYear{2022}

\newcommand{\ourModelName}{ShadingNet}

\newcommand{\ourRepName}{RiCS}

\newcommand{\ourTitle}{\ourRepName: A 2D Self-Occlusion Map for Harmonizing Volumetric Objects}

\begin{document}

\title{\ourTitle}

\author{\bf Yunseok Jang\thanks{Major part of this work has been done while Yunseok Jang was an intern at Adobe Research.}~$^{\clubsuit}$ \: \:
  Ruben Villegas$^{\spadesuit}$ \: \:
  Jimei Yang$^{\diamondsuit}$ \: \: %
  Duygu Ceylan$^{\diamondsuit}$ \: \:
  Xin Sun$^{\diamondsuit}$ \: \:
  Honglak Lee$^{\clubsuit}$ \\
  $^\clubsuit$University of Michigan, Ann Arbor,~~$^\spadesuit$Google Brain,~~$^\diamondsuit$Adobe Research\\
  }

\maketitle

\begin{abstract}

There have been remarkable successes in computer vision with deep learning.
While such breakthroughs show robust performance, there have still been many challenges in learning in-depth knowledge, like occlusion or predicting physical interactions.
Although some recent works show the potential of 3D data in serving such context, it is unclear how we efficiently provide 3D input to the 2D models due to the misalignment in dimensionality between 2D and 3D.
To leverage the successes of 2D models in predicting self-occlusions, we design Ray-marching in Camera Space (\ourRepName), a new method to represent the self-occlusions of foreground objects in 3D into a 2D self-occlusion map.
We test the effectiveness of our representation on the human image harmonization task by predicting shading that is coherent with a given background image.
Our experiments demonstrate that our representation map not only allows us to enhance the image quality but also to model temporally coherent complex shadow effects compared with the simulation-to-real and harmonization methods, both quantitatively and qualitatively. 
We further show that we can significantly improve the performance of human parts segmentation networks trained on existing synthetic datasets by enhancing the harmonization quality with our method.

\end{abstract}

\section{Introduction}
\label{sec:introduction}

\begin{figure}[t]
    \centering
    \includegraphics[width=\linewidth]{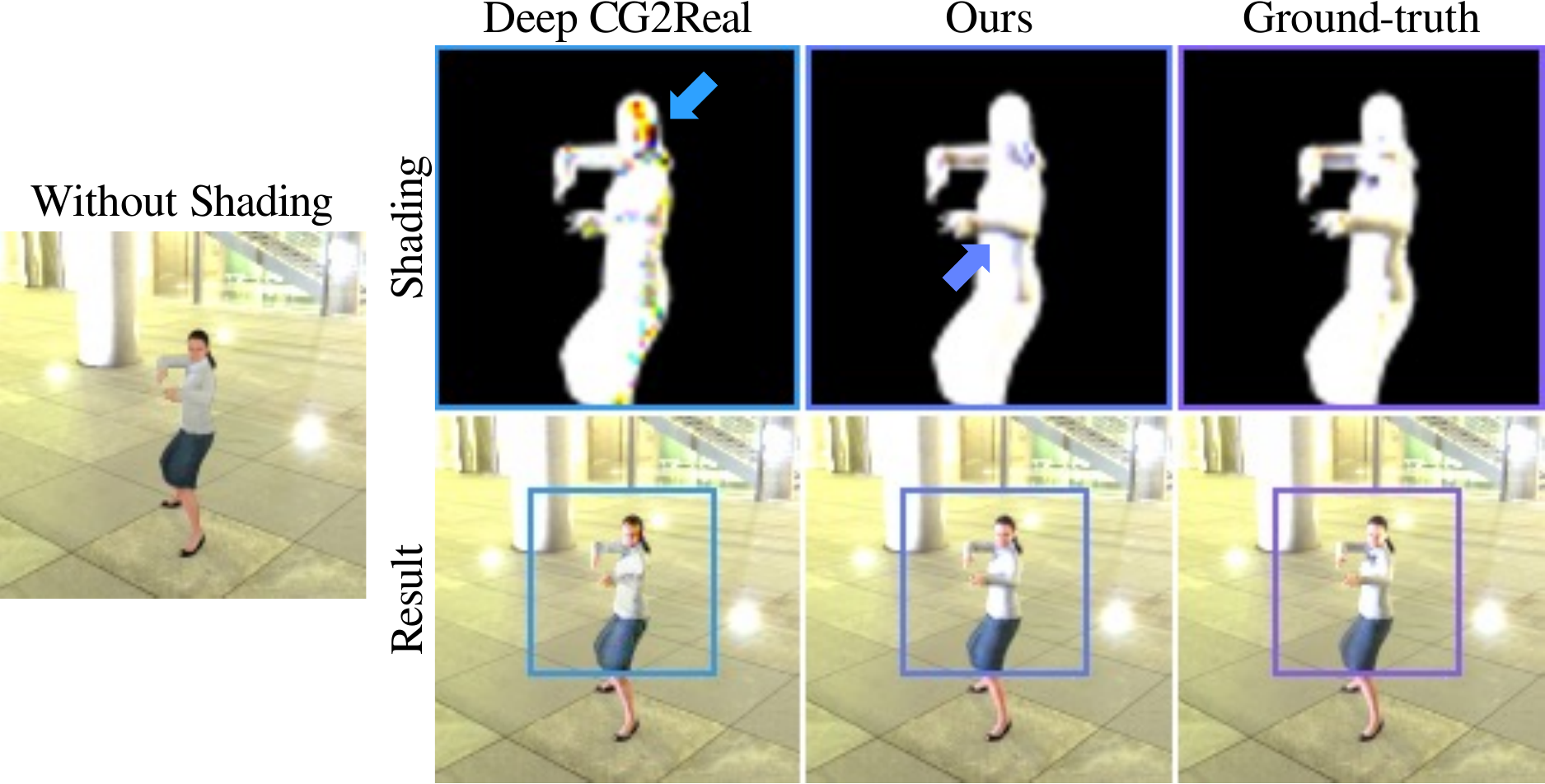}
    \caption{We propose~\ourModelName, which is based on~\ourRepName, our novel 2D~\emph{self-occlusion}~map that enables understanding volumetric objects without taking any 3D inputs. %
    Based on the robust~\ourRepName~\emph{self-occlusion}~map,~\ourModelName~better harmonizes the foreground human with fewer noises around the head and richer shadows around the arms than Deep CG2Real~\cite{bi-iccv19}.
    }
    \label{fig:shading_map_visual_results}
    \vspace{-6pt}
\end{figure}

The ability to predict occlusions, volume, or physical behaviors from a single image relies on a strong contextual knowledge learned from experience. Although breakthroughs in 2D image understanding have been made within the last decade, such as classification~\cite{he-cvpr16,simonyan-iclr15}, detection~\cite{noh-cvpr18,redmon-cvpr16,ren-neurips15,singh-iccv17}, segmentation~\cite{he-iccv17,noh-iccv15}, mining volumetric information still remains as a challenge.
To date, various works contributed to address those problems, from intentionally masking an image~\cite{noh-cvpr18,singh-iccv17} to handling 3D inputs~\cite{liu-neurips20,mildenhall-eccv20}. %
While augmentations~\cite{noh-cvpr18,singh-iccv17} could help a model learn more general patterns from data, it does not provide explicit volumetric information.
On the other hand, 3D inputs could provide contextual details, but understanding 3D data presents a bigger challenge compared to 2D images.%

In this context, we propose to encode the volumetric context of an object in a 2D shape so that we can provide appropriate contextual knowledge, especially about self-occlusion, without increasing the dimensionality of the input data. 
With this goal in mind, we introduce Ray-marching in Camera Space (\ourRepName), a method for recording \textit{self-occlusions} in a 2D representation map of 3D objects. 
Since we directly obtain our~\ourRepName~map from a volumetric object instead of predicting it from a neural network, our representation could work as a robust context provider to any learnable component.

We apply \ourRepName, our self-occlusion representation, to the moving object setting to validate the robustness of our approach.
Among various applications, we focus on predicting the shadings of various humans given a background image in the Image Harmonization~\cite{cong-cvpr20,pitie-iccv05,reinhard-cgna01,sunkavalli-tog10,tsai-cvpr17} task setting, based on the idea that we naturally learn the effect of occlusion in the real world by watching the shadows.
To effectively handle our~\ourRepName~representation for this task, we design a model named~\ourModelName.
Additionally, we introduce a proxy loss so that~\ourModelName~can predict complex shadow effects generated by the relative position of the human and the light source in the background. %

For evaluating the effectiveness of our~\ourRepName~map, we test our approach on a simulated dataset, which contains a total of 212,955 frames from 38 humans with 398 motion sequences.
Compared to the image-to-image translation~\cite{park-eccv20,wang-cvpr18}, harmonization~\cite{cong-cvpr20}, and simulation-to-real~\cite{bi-iccv19}, our method generates better shading, not just per-frame but also in a temporally coherent manner.
In addition to it, we also demonstrate the effectiveness of our approach in training a neural network by applying our shading prediction to SURREAL~\cite{varol-cvpr17}, an existing large-scale synthetic human dataset.
We show that the neural network model trained on a dataset with more harmonized shadings significantly outperforms the model trained on the original dataset and generalizes better to the real data.

The contributions of our work are summarized as follows:
\begin{compactitem}
    \item We introduce a novel representation that models self-occlusion of a volumetric object in a 2D map. %
    \item We propose a neural network-based approach that generates realistic shading of a foreground human that harmonizes well with a given background image.
    \item Our method gives improved shading prediction results compared with the image harmonization and synthetic-to-real approaches.
    \item We apply our shadings to an existing synthetic dataset pipeline and show better generalization ability on human parts segmentation networks.
\end{compactitem}

\section{Related Work}
\label{sec:related-work}
\textbf{Image-to-Image Translation.} %
Image-to-image translation for realistic image synthesis has been extensively studied~\cite{liu-neurips17,park-eccv20,wang-cvpr18,zhu-iccv17,zhu-neurips17}.
Isola~\etal~\cite{isola-cvpr17} perform paired image-to-image translation using an adversarial loss~\cite{goodfellow-neurips14}, with a regression loss between the generated result and the target.
Along with this work, previous studies~\cite{bi-iccv19,wang-cvpr18} apply regression not only on RGB image space but also in perceptual feature space~\cite{johnson-eccv16}.
Recently, some approaches, including Park~\etal~\cite{park-eccv20}, start addressing an unpaired setting by adding a contrastive loss~\cite{oord-arxiv18}.
However, all these methods may introduce visual artifacts in the generated images, especially when finding relationships between the two domains is difficult.

Among all previous studies, Deep CG2Real~\cite{bi-iccv19} tries to learn a translation between the OpenGL rendering and realistic images.
It performs a two-stage generation process by first synthesizing a Physically-Based Renderer (PBR) image of the full image scene using albedo and shading maps and then translating the PBR image into a more realistic image.
However, in the real world, it is challenging to obtain the ground-truth albedo and shading map from a background image.
Thus, we address this problem by predicting the light representation from the background image instead.
Also, to generate a detailed shading of a complex-shaped object (\ie, human), we introduce~\ourRepName, a novel way to encode self-occlusion in a 2D form.
These two approaches allow the model to learn the relationship between the two domains more clearly, thereby generating a more precise harmonization output. 
Conceptually, all methods in this section could be used to translate synthetic humans into humans that are harmonized with a background image. 
Nevertheless, all the aforementioned works do not explicitly model how light hits the geometry of the foreground human.
Correct light and volumetric modeling of humans are critical to accurately harmonize synthetic humans into realistic videos.

\textbf{Deep Image Harmonization.}
Modern graphics engine still uses Monte Carlo Rendering as an approximation because it is infeasible to exactly predict the rays.
However, if the light source is dominated by a few strong sources like the sun, we could predict the general shadings.
In this context, earlier works~\cite{hold-cvpr17,zhang-cvpr19} focused on predicting the sunlight position, assuming that the primary light source of an outdoor image is the sunlight.

Meanwhile, image harmonization methods try to avoid low-level appearance statistics matching~\cite{pitie-iccv05,reinhard-cgna01,sunkavalli-tog10}.
Instead, recent works start to understand the geometry and lighting details from the background. 
For instance, Tsai~\etal~\cite{tsai-cvpr17} first harmonize the foreground object with the background by obtaining global information of the background from an end-to-end network.
Following Tsai~\etal~\cite{tsai-cvpr17}, Cong~\etal~\cite{cong-cvpr20} introduce a domain verification discriminator to improve the quality of image harmonization.
However, because both Tsai~\etal~\cite{tsai-cvpr17}~and Cong~\etal~\cite{cong-cvpr20} do not take any information about geometry in their problem settings, it is hard to harmonize the foreground object when the surface is complex-shaped (\eg~a human body).
To avoid such an issue, we provide the surface normal map and \emph{self-occlusion} representation which provides more volumetric context for better shading prediction.

\textbf{Knowledge Transfer from Simulation to Real.}
There have been extensive works in guiding neural networks to perform better on real tasks by providing simulated data.
For example, FlowNet~\cite{dosovitskiy-iccv15} learns to estimate the optical flow by synthetic 3D moving chairs.
Like what we predict the shading over simulated humans, there has been a lot of works in adopting simulations: 2D pose~\cite{pishchulin-cvpr12}, 3D pose~\cite{sminchisescu-cvpr06}, action recognition~\cite{rahmani-cvpr15}, etc.
In this context, SURREAL~\cite{varol-cvpr17} is one of the representative works that generate large-scale data by employing the human motion, body model, and spherical harmonics with nine coefficients~\cite{green-gdc03}.
Based on its large-scale synthetic data, SURREAL presents performance improvement in part segmentation and depth prediction tasks.

Although various works try to employ simulation data to improve performance in real tasks, models trained with synthetic images do not always perform better on real images due to the gap in the data distribution~\cite{bi-iccv19}.
To reduce such gap, some of the domain adaptation work tries to add task-specific constraints like the target should be closer to the source~\cite{mueller-cvpr18} or introduce a task-specific loss like preserving cycle consistency between the two domains~\cite{hoffman-icml18}.
To validate the effectiveness in downstream tasks, we follow the same pipeline as SURREAL~\cite{varol-cvpr17} but use our method to render higher quality data. %

\section{Method}
\label{sec:methods}
\begin{figure}[t]
    \centering
    \includegraphics[width=\linewidth]{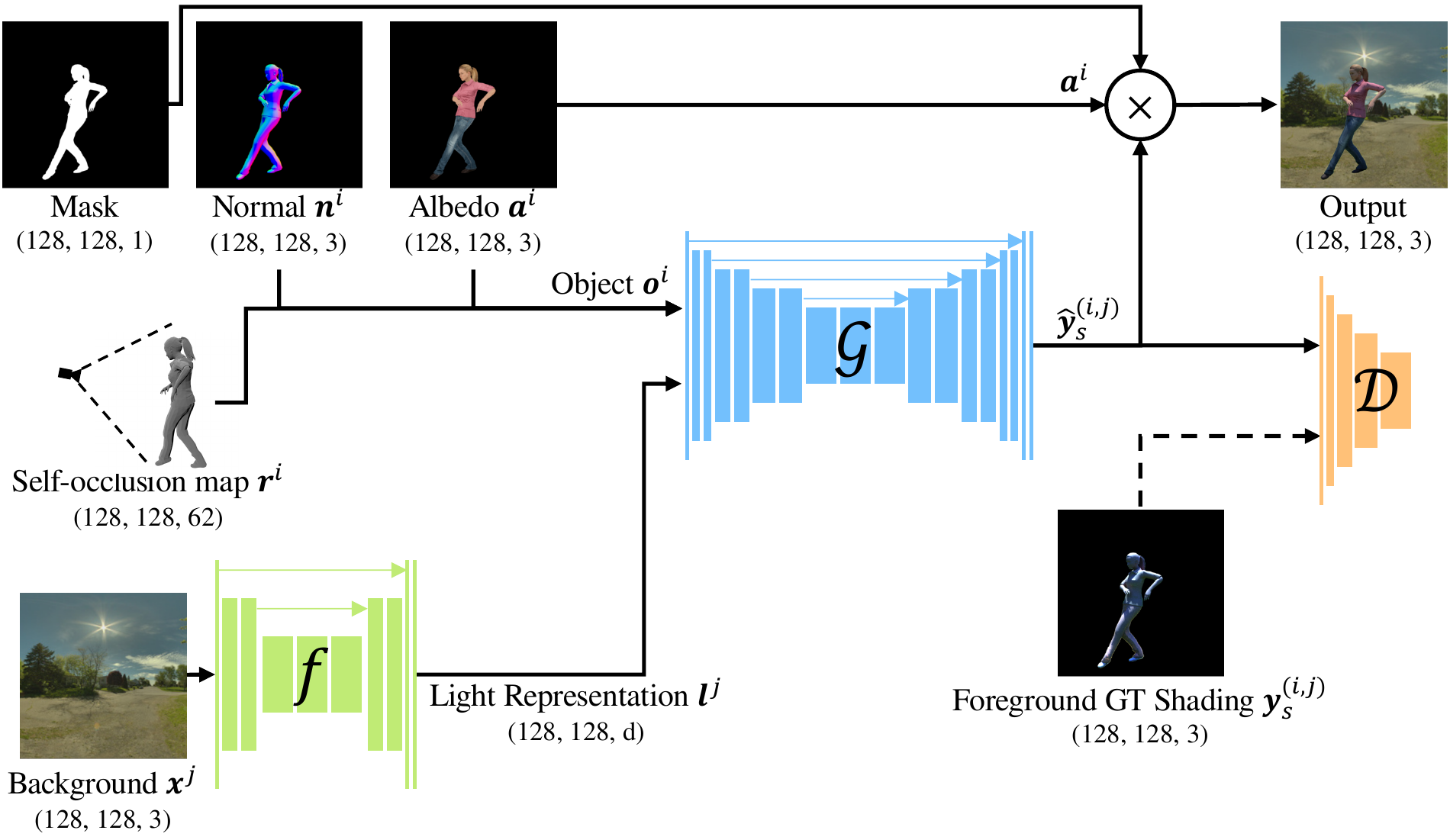}
    \caption{\textbf{Bird-eye view of our~\ourModelName.} Our goal is to generate visual output from the foreground object~$\mathbf{o}^{i}$, including our~\emph{2D self-occlusion}~map (\ourRepName), and a light representation~$\mathbf{l}^{j}$~predicted from the background~$\mathbf{x}^{j}$. }
    \label{fig:birdeye_view}
    \vspace{-6pt}
\end{figure}

Given a background image and a foreground rendering of a human, our goal is to synthesize realistic shading for the human that enables harmonizing the foreground and the background.
In this section, we describe our end-to-end shading synthesis approach, named~\ourModelName, which consists of four main components.

First of all, we design a \emph{deep light encoder}~($f$ in Fig.~\ref{fig:birdeye_view}) that extracts a light source representation from the background image.
Then, we introduce Ray-marching in Camera Space (\ourRepName), a novel \emph{self-occlusion map}~($\mathbf{r}^{i}$ in Fig.~\ref{fig:birdeye_view}) which provides self-occlusion information in matched camera space as a 2D representation.
Once we get the 2D representations from previous steps, we feed those contextual information to a \emph{deep shading generator}~($\mathcal{G}$ in Fig.~\ref{fig:birdeye_view}) that synthesizes plausible and realistic shading.
Finally, we utilize a discriminator~($\mathcal{D}$ in Fig.~\ref{fig:birdeye_view}) to guide the shading generation network to generate more realistic outputs.

\subsection{Deep Light Encoder}
\label{subsec:deep_light_encoder}
Image-Based Lighting (IBL) is a commonly used light representation in 3D rendering, where the real-world light information is projected into a 2D image.
Hence, reconstructing the exact light representation is a challenging problem because it needs to understand a 3D lighting condition from a single 2D image, which often includes complex light interactions.
To tackle this challenge, we split the role of light representation into light sources and interaction with the objects.

To extract the details about the light source from the background image at test time, we employ a light encoder $f$.
However, because there is no label between the light source and the light interaction of an object, we use a synthetic dataset generated by a physically-based renderer at training time.
So, for each rendered background image $\mathbf{x}^{j}$, we have access to the ground truth IBL configuration and the $4\times4$ camera matrix parameters used to render the image.
With the IBL and the camera matrix, we can focus primarily on predicting light sources from the 2D background image $\mathbf{l}^{j} = f(\mathbf{x}^{j})$ without any human-annotated labels.
We refer to Appendix~\ref{supp_subsec:light_encoder}~for the detailed architectures of the network $f$.

\begin{figure}[t]
    \centering
    \includegraphics[width=\linewidth]{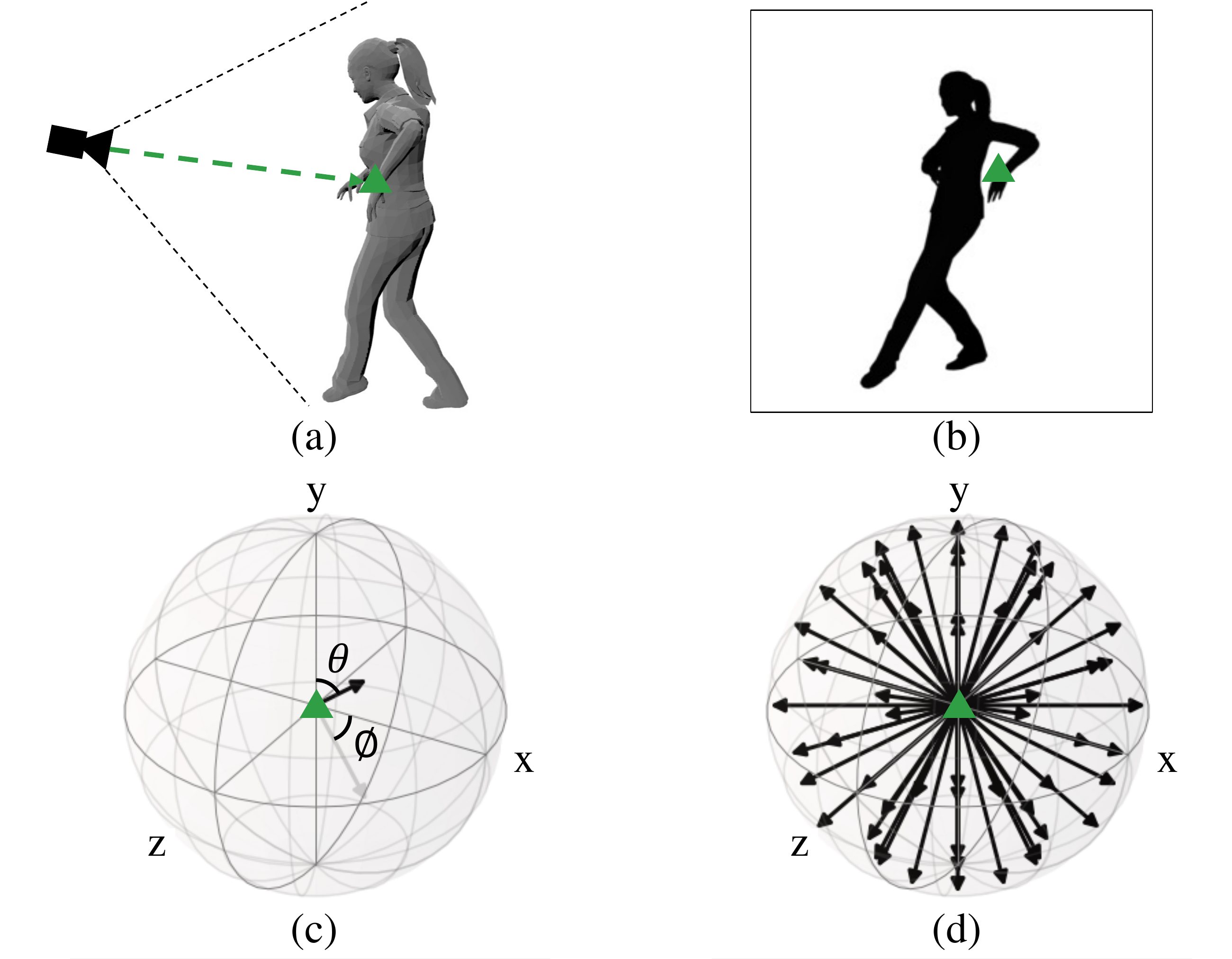}
    \caption{\textbf{Visualization of the steps in generating~\ourRepName, our \emph{self-occlusion representation} $\mathbf{r}$.} We first (a) ray-marching (width x height) rays in camera space. As a result, (b) we discover at most (width x height) points, marked as black, that lie on the face of an object, like the green triangle around the left wrist in Figures (a) and (b). Then, we shoot 62 rays, as visualized in (d), for every point discovered from (a). We record whether it hits the 3D object or not as a binary value and use it as $\mathbf{r}$.
    We sample rays by rotating $\phi$ and $\theta$ in (c) every 30 degrees in camera space. Each ray is solely used for checking whether it hits the foreground surface. %
    }
    \label{fig:occlusion_ray_visualize}
    \vspace{-6pt}
\end{figure}

\subsection{\ourRepName: A Novel 2D Self-Occlusion Map}
\label{subsec:rocs_occlusion_map}
While the albedo~$(\mathbf{a}^{i})$~and normal~$(\mathbf{n}^{i})$ map could provide a piece of global information about a foreground human, it is often insufficient to predict certain local and complex shading effects such as self-shadows that occur commonly as a human is moving.
Therefore, we introduce a method for representing potential \emph{self-occlusions}~in 2D: Ray-marching in Camera Space (\ourRepName).

Precisely, as visualized in (a) and (b) of Fig.~\ref{fig:occlusion_ray_visualize}, we first discover the points in 3D space by ray-marching to each 2D pixel location of the final output (height $h$ $\times$ width $w$) from the camera position, which is given as a $4\times4$ camera matrix.
In theory, this step, similar to the 2D projection, can discover up to height $\times$ width 3D points.
However, practically, it discovers fewer points because not all pixels are covered by the foreground object in most cases.

Then, for each 3D point that lies on the surface of a foreground object, we shoot a fixed set of rays sampled uniformly at equal angle intervals, as in Figs.~\ref{fig:occlusion_ray_visualize}.(c) and \ref{fig:occlusion_ray_visualize}.(d).
For the efficiency, we set 30 degrees as sample frequency to store the 62 bits to a single 64-bit integer.
We record whether each ray hits the foreground object or not as a boolean vector $\mathbf{r}^{i} \in \mathds{R}^{h \times w \times 62}$, where a discovered hit corresponds to a self-occlusion case.
We would like to note that we are not penetrating nor reflecting rays but record whether each sampled ray hit the foreground surface or not.

Our~\ourRepName~map is assumed to be fed into a series of 2D convolution operations in a pixel-aligned form, which is a common use case in computer vision~\cite{he-cvpr16,ronneberger-miccai15}.
By comparing the difference to the surrounding pixels, we expect the network to estimate rich self-occlusion information.
Likewise, once a model is aware of the occluded directions for each pixel position, we believe the model could better describe which light sources are affected by the self-occlusions.
For the rest of this paper, we refer to this representation as the 2D \emph{self-occlusion}~\ourRepName~map $\mathbf{r}^{i}$ and assume our~\ourRepName~map is precomputed before training a network.

\subsection{Deep Shading Generator}
\label{subsec:deep_shading_generator}

We define a U-Net~\cite{ronneberger-miccai15} based generator~\cite{cong-cvpr20,zhu-neurips17} $\mathcal{G}$ that takes 2D representations of the object $\mathbf{o}^{i} = (\mathbf{n}^{i}, \mathbf{a}^{i}, \mathbf{r}^{i})$ consisting of the normal ($\mathbf{n}^{i}$), albedo ($\mathbf{a}^{i}$), and our novel \emph{self-occlusion} \ourRepName~map ($\mathbf{r}^{i}$) rendered from the desired viewpoint, along with the light embedding $\mathbf{l}^{j}$ as pixel-aligned 2D input.
The output of generator $\mathbf{y}_{s}^{(i,j)} = \mathcal{G}(\mathbf{o}^{i}, \mathbf{l}^{j})$ is a shading map.
Our U-Net-based generator is composed of four Downscale-Upscale blocks with a linear output layer (see Appendix~\ref{supp_subsec:shading_generator}~for details).
Given the normal, albedo, and our~\ourRepName~map from a foreground object, the background image, and a foreground mask, we can synthesize the harmonized output image as shown in Fig.~\ref{fig:birdeye_view}.

While our~\ourRepName~map helps identify the regions in the image where self-shadowing should occur, these regions are often relatively small and hence can be ignored by the network.
To focus the attention of the network on such regions without any explicit labels, we employ the Canny Edge Detection~\cite{canny-ieee86}.
It is based on an intuition that we can easily detect bright-to-dark color changes in the shading map caused by self-occlusions of body parts as edges.
Based on such an idea, we employ the computed edges from the foreground shading as the regression loss's importance $\textbf{x}_{w}$ for each pixel.

Given the predicted shading map $\mathbf{\hat{y}}_{s}^{(i,j)} = \mathcal{G}( \mathbf{o}^{i}, \mathbf{l}^{j})$ and the importance weights $\textbf{x}_{w}$, we apply an $\ell_{1}$ regression loss against the ground-truth shading map $\mathbf{y}_{s}^{(i,j)}$:
\begin{align}
\label{eq:regression_shading}
    \mathcal{L}^{\text{recon}}_{s} = \textbf{x}_{w}~\odot~\lVert\mathbf{\hat{y}}_{s}^{(i,j)} - \mathbf{y}_{s}^{(i,j)}\lVert_{1},
\end{align}
where $\odot$ means the element-wise product, and $\lVert\cdot\lVert_{1}$ stands for $\ell_{1}$-distance.
Finally, we apply an additional loss $\mathcal{L}^{\text{recon}}_{g} = ||\mathbf{\hat{y}}_{g}^{(i,j)} - \mathbf{y}_{g}^{(i,j)}||_{1}$ on the final RGB image rendered by our model $\mathbf{\hat{y}}_{g}^{(i,j)} = \mathbf{\hat{y}}_{s}^{(i,j)}~\odot~\mathbf{a}^{i}$, where $\mathbf{a}^{i}$ is the human albedo. We do not use importance weighing on the RGB image.

\subsection{A Conditional Discriminator}
\label{subsec:conditional_discriminators}

To further improve the realism of the shading map produced by our generator, we design a conditional discriminator $\mathcal{D}_{(s,l)}$ that helps our generator better fit its output $\mathbf{\hat{y}}_{s}^{(i,j)}$ to the estimated light $\mathbf{l}^{j}$ and input conditions $(\mathbf{n}^{i}, \mathbf{a}^{i}, \mathbf{r}^{i}) = \mathbf{o}^{i}$.
Our discriminator is implemented similar to the original GAN~\cite{goodfellow-neurips14} but with the label injection technique~\cite{miyato-iclr18} to better model the conditions:
\begin{align}
\label{eq:condgan_loss}
\nonumber \mathcal{L}^{\text{GAN}}_{s} \left( \theta_{\mathcal{G}}, \theta_{\mathcal{D}} \right) =~ &\mathbb{E}_{\mathbf{y}_{s} \sim p_{\text{data}}\left( \mathbf{y}_{s} \right)}\left[ \log \mathcal{D} \left( \mathbf{y}_{s}^{(i,j)}|\left( \mathbf{o}^{i}, \mathbf{l}^{j} \right) \right) \right] \\
+ \mathbb{E}_{\mathbf{\hat{y}}_{s} \sim p_{\text{gen}}\left(\mathbf{\hat{y}}_{s}\right)}&\left[ \log \left( 1 - \mathcal{D} \left( \mathbf{\hat{y}}_{s}^{(i,j)}|\left( \mathbf{o}^{i}, \mathbf{l}^{j} \right) \right) \right) \right]
\end{align}
where $\mathbf{\hat{y}}_{s}^{(i,j)} = \mathcal{G}( \mathbf{o}^{i}, \mathbf{l}^{j})$.
Also, we train our discriminator to capture more details by modeling the shading map at the patch level using PatchGAN~\cite{isola-cvpr17}.
Please check the detailed architecture in Appendix~\ref{supp_subsec:shading_discriminator}.

Lastly, to stabilize the training dynamics, we employ a feature matching loss from the discriminator, following Pix2PixHD~\cite{wang-cvpr18}. The predicted shading $\mathbf{\hat{y}}_{s}$ is compared with the ground-truth $\mathbf{y}_{s}$ at multiple layers of the discriminator.
Denoting the $k^{\text{th}}$~layer feature, from the input, of the discriminator $D$ as $D^{k}$, then feature matching loss $L^{\text{FM}}_{s}$ is defined as:
\begin{align}
\label{eq:feature_matching}
    \mathcal{L}^{\text{FM}}_{s} = \mathbb{E}_{(\mathbf{y}_{s}, \mathbf{\hat{y}}_{s})} \sum_{k} \frac{1}{N^{k}} \lVert
    \mathcal{D}^{k} \left( \mathbf{y}_{s}|\left( \mathbf{o}, \mathbf{l} \right) \right)
    - \mathcal{D}^{k} \left( \mathbf{\hat{y}}_{s}|\left(\mathbf{o}, \mathbf{l} \right)\right)\lVert_{1},
\end{align}
where $N^{k}$ is the number of elements in the $k^{\text{th}}$~layer.

\subsection{Training Objective}
\label{subsec:training_losses}

\begin{figure*}[t]
    \centering
    \includegraphics[width=\linewidth]{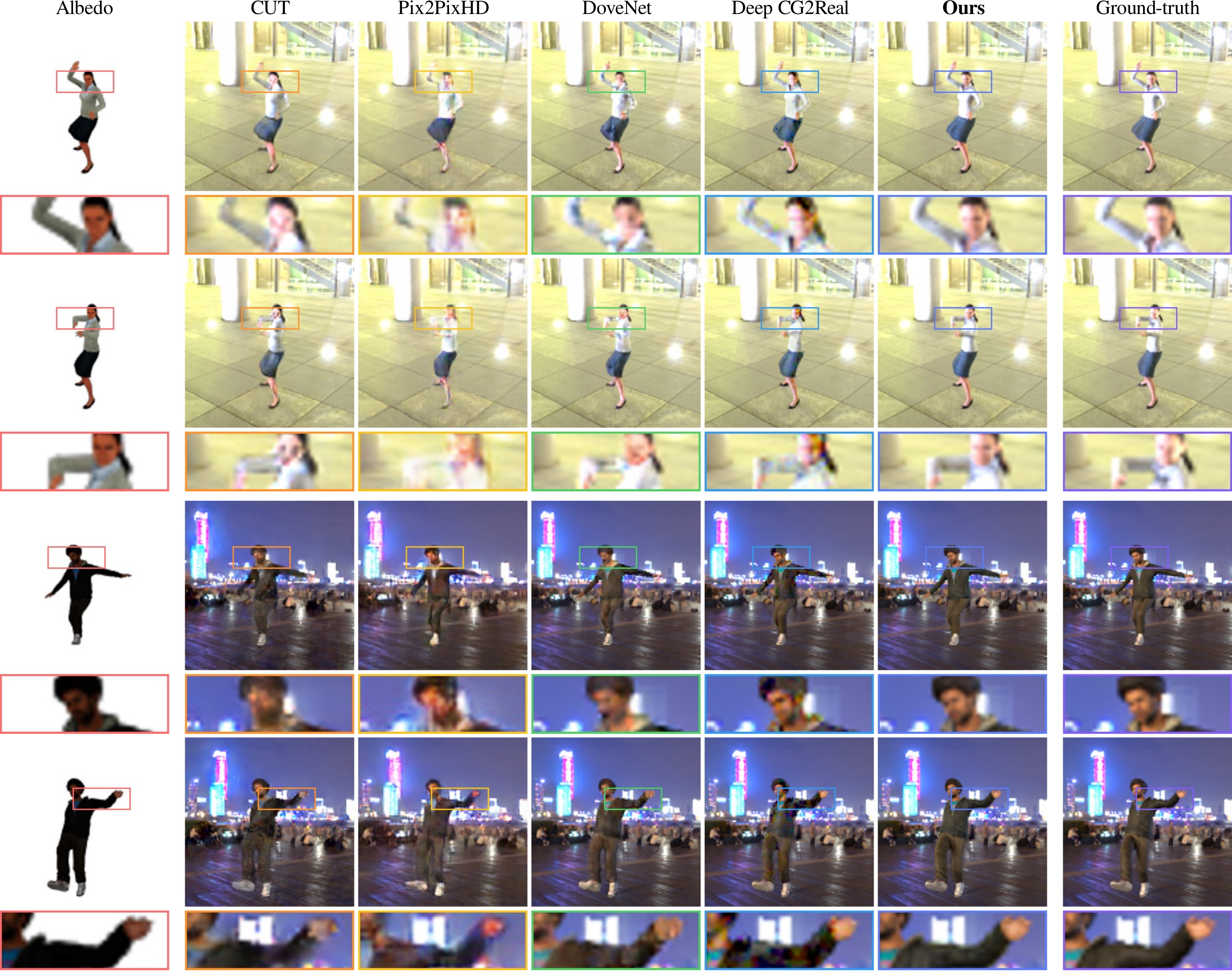}
    \caption{\textbf{Shading generation result.} All networks generate the shading of a human from a background.
    Each row comes with an upscaled rectangle from the marked position for better comparison.
    The upper two and lower two rows come from the same test sequence.
    Ours shows significantly fewer pixel noises on the face and arms.
    In-depth analysis of the second row can also be found in Figure~\ref{fig:shading_map_visual_results}.
    }
    \label{fig:qualitative_baseline_comparison}
\end{figure*}

Our network is trained with three types of losses.
First, we train our shading map generator using an edge-guided shading map regression loss~$\mathcal{L}^{\text{recon}}_{s}$ in Eq.~\ref{eq:regression_shading}, along with an RGB regression loss~$\mathcal{L}^{\text{recon}}_{g}$.
Second, we employ a discriminator that provides a condition mismatching signal to the generator via the objective $\mathcal{L}^{\text{GAN}}_{s}$ (Eq.~\ref{eq:condgan_loss}).
Third, to stabilize the training dynamics, we introduce a feature matching loss~$\mathcal{L}^{\text{FM}}_{s}$ in Eq.~\ref{eq:feature_matching}.
Overall, our training objective is defined by
\begin{align}
\label{eq:loss_all}
    \nonumber \min_{\theta_{\mathcal{G}}} \max_{\theta_{\mathcal{D}}} ~&
    \mathcal{L}^{\text{GAN}}_{s} \left( \theta_{\mathcal{G}}, \theta_{\mathcal{D}} \right) + \mathcal{L}^{\text{FM}}_{s} \left( \theta_{\mathcal{G}} \right) \\
    & + \mathcal{L}^{\text{recon}}_{s} \left( \theta_{\mathcal{G}} \right) + \mathcal{L}^{\text{recon}}_{g} \left( \theta_{\mathcal{G}} \right),
\end{align}
where each $\mathcal{L}^{\left\{\cdot\right\}}$ has its own loss weight parameter to balance its influence (see Appendix~\ref{supp_sec:training}).

\section{Experiments}
\label{sec:experiments}
We evaluate our approach in the following settings. 
First, we compare image harmonization results with the baselines.
Next, we use our method to synthesize data to improve the performance of off-the-shelf segmentation estimation methods on real images.

\subsection{Comparison on Visual Quality}
\label{subsec:baseline_comparison}

In this section, we qualitatively and quantitatively compare the result of our~\ourModelName~method against the baselines in harmonizing the foreground object with the background. 
We first introduce the dataset and the baselines we use and then discuss the main results.

\textbf{Dataset.}
For training and testing our method as well as the baselines, we construct a large-scale dataset by animating and rendering synthetic humans.
Specifically, we retarget 398 randomly selected motion sequences from~Mixamo~\cite{adobe-mixamo} into 38 3D human models from RocketBox~\cite{gonzalez-fvr20}.
We render images with three evenly distributed cameras in front of the human body for each animated sequence. %
Each camera is set to track the character's chest in the center, and its trajectory is further smoothed to avoid abrupt camera motion. 
The camera distance is also slightly perturbed to bring more diversity. 
In total, our dataset consists of 212,955 frames.
We use the rendered images from 30 characters for training and images from the remaining eight characters for testing. 
For each rendered image $\mathbf{y}_{g}$, we obtain the 3D body mesh, the camera parameters, albedo, surface normal recorded in camera space, and foreground alpha mask.
Please see Appendix~\ref{supp_sec:dataset} for the details.

\textbf{Implementation Details.}
We trained~\ourModelName~for 200 epochs, with a batch size of 54 images per GPU.
Also, we use a mixed-precision~\cite{micikevicius-iclr18} `opt\_level = O1' setting on 8 Titan Xp GPUs on Ubuntu 20.04 LTS machine with PyTorch 1.5.1, CUDA 10.2, and cuDNN 7.6.
We set automatic learning rate scaling and optimized with the fused version of ADAM~\cite{kingma-iclr15} optimizer, named FusedAdam, in the APEX~\cite{apex-20} library.
More details about implementation and training are in Appendix~\ref{supp_sec:training}.

\textbf{Baselines.}
We compare our method with four baselines from the computer vision domain that are not taking 3D inputs: CUT~\cite{park-eccv20} (unpaired image transfer), Pix2PixHD~\cite{wang-cvpr18} (paired image transfer), DoveNet~\cite{cong-cvpr20} (image harmonization), and the OpenGL to PBR transfer introduced in Deep CG2Real~\cite{bi-iccv19}. 
For CUT and DoveNet, we provide the human albedo map overlaid on the background as input and generate the harmonized composition. 
For Pix2PixHD and Deep CG2Real, we provide the albedo, normal, alpha, and background images as separate 2D inputs to synthesize the foreground human.
We train each method on 128$\times$128 images for 200 epochs. 
We experimented with training the baseline methods on single frames, but we observed that they did not perform well. 
Therefore, we feed three frames from three different viewpoints, without any augmentations, as input to all models for a more fair comparison.
To the best of our knowledge, no prior work has attempted to solve the self-occlusion issue in Image Harmonization using 3D operations.
Please check Appendix~\ref{supp_sec:additional_experiments}~for additional comparisons.

\textbf{Results.}
We first conduct evaluations with five different metrics over the rendered foreground humans. %
We employ Mean Squared Error (MSE) for directly measuring the distance in RGB space.
Then, to measure the visual quality, we additionally use Learned Perceptual Image Patch Similarity (LPIPS)~\cite{zhang-cvpr18} and Structural Similarity (SSIM)~\cite{wang-tip04}, which have been used as the standard in computer vision as an image quality measure.
To measure the temporal coherence of the sequences generated by each method, we employ Fr\'echet Video Distance (FVD)~\cite{unterthiner-iclrw19} as a video feature distribution distance and Motion-based Video Integrity Evaluation (MOVIE)~\cite{seshadrinathan-tip09} as a video distortion measure.
Because MOVIE is defined for more than 33 frames, but we feed 30 frames per each test sequence, we pad the first and the last frame two times.

Table~\ref{tab:quantitative_baseline_comparison} summarizes the quantitative results from the visual quality metrics among different network models.
Our method generates higher quality output than other baselines in all image distance (42.77\% lower in MSE), image quality (41.78\% lower in LPIPS), and temporal stability (49.34\% lower in FVD). 
To better address such differences, we perform a qualitative comparison in Figs.~\ref{fig:shading_map_visual_results}~and~\ref{fig:qualitative_baseline_comparison}.

\begin{table}[t]
\setlength{\tabcolsep}{2pt}
\small
\centering
\begin{tabular}{l|*5c}
\toprule
\multirow{2}{*}{Method}       & MSE$\downarrow$ & LPIPS$\downarrow$ & SSIM $\uparrow$ & FVD$\downarrow$ & MOVIE$\downarrow$ \\
                              & {\scriptsize($\times 10^{-3}$)} & {\scriptsize($\times 10^{-2}$)} & {\scriptsize($\times 10^{-1}$)} & {\scriptsize($\times 10^{1}$)} & {\scriptsize($\times 10^{-4}$)} \\
\midrule
CUT~\cite{park-eccv20}        & 11.68 & 2.93 & 9.68 & 9.99 & 22.39 \\
Pix2PixHD~\cite{wang-cvpr18}  &  9.89 & 3.01 & 9.65 & 9.10 & 17.10 \\
DoveNet~\cite{cong-cvpr20}    &  7.72 & 2.13 & 9.72 & 12.47 & 19.48 \\
Deep CG2Real~\cite{bi-iccv19} &  6.50 & 2.19 & 9.71 & 10.11 & 11.10 \\
\textbf{\ourModelName~(full)} & \underline{3.72}  & \textbf{1.24} & \textbf{9.87} & \underline{4.61} & \textbf{7.00} \\
\midrule
No $\mathcal{L}^{\text{FM}}_{s}$    &  4.02 & \underline{1.30} & 9.86 & 5.00 & 7.68 \\
No $\mathcal{L}^{\text{recon}}_{s}$ &  \textbf{3.58} & 1.32 & \underline{9.86} & 5.71 & \underline{7.52} \\
No $\mathcal{L}^{\text{recon}}_{g}$ &  4.89 & 1.92 & 9.84 & 9.42 & 8.33 \\
No $\mathbf{r}^{i}$                 &  4.65 & 1.54 & 9.84 & \textbf{3.67} & 8.67 \\
\bottomrule
\end{tabular}
\caption{\textbf{Quantitative result on generated image quality.} We measure the visual quality between the output from a physically-based renderer and each method. We also perform an ablation study by removing one of the main components from our full network and mark each model as `No X'.~$\uparrow$ next to the evaluation metric means a higher number is favored for that metric, and $\downarrow$ is for the opposite.} 
\label{tab:quantitative_baseline_comparison}
\end{table}

\begin{figure*}[t]
    \centering
    \includegraphics[width=\linewidth]{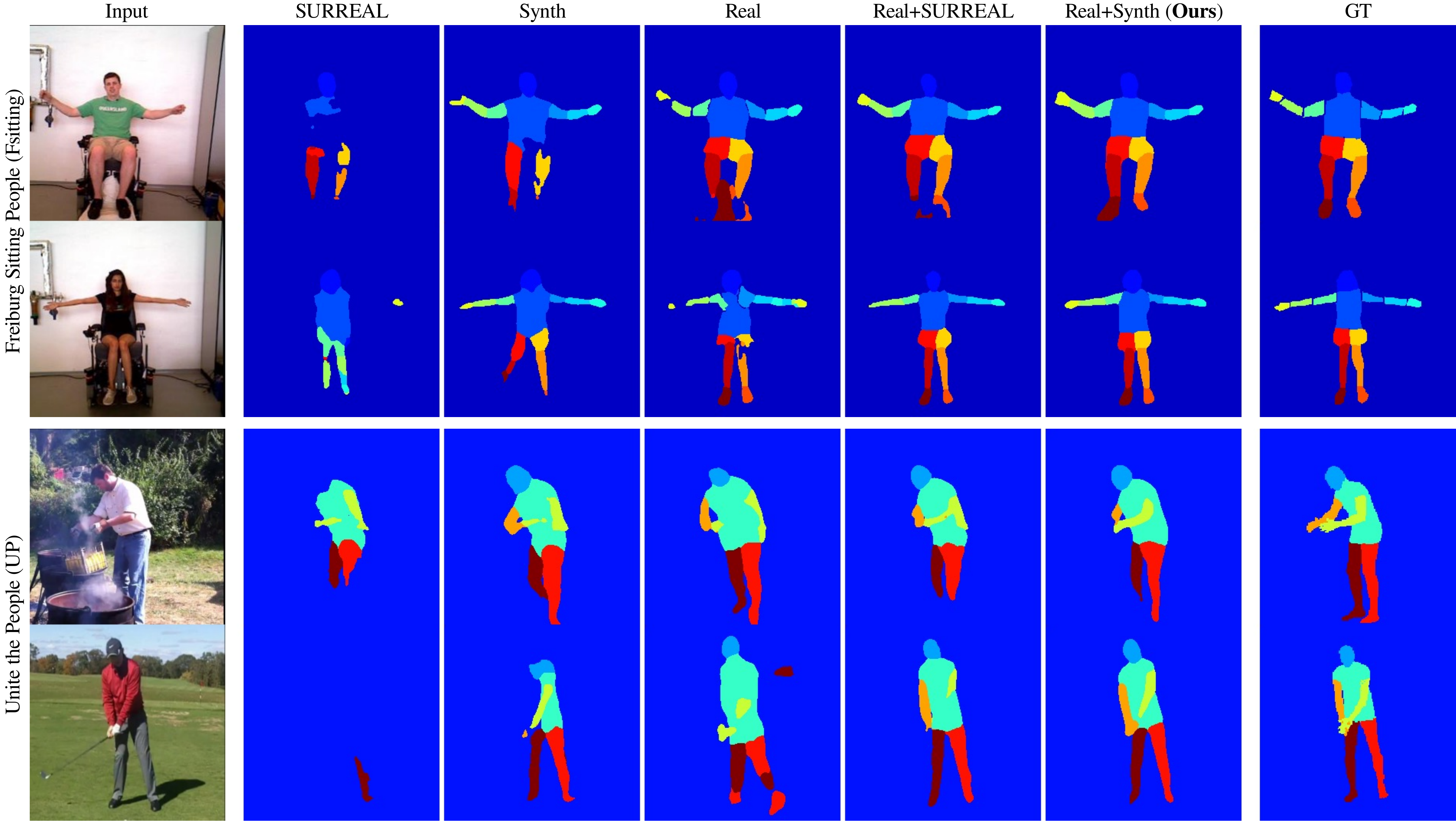}
    \caption{\textbf{Part segmentation predictions on Freiburg Sitting People~\cite{oliveira-icra16} and Unite the People~\cite{lassner-cvpr17} dataset.} 
    SURREAL stands for the model trained with the SURREAL dataset~\cite{varol-cvpr17}, and Synth stands for the model trained with the data which includes our shadings.
    \textbf{Ours} gives better segmentation (\eg~arms and legs in each example).
    }
    \label{fig:part_segmentation_visual_results}
\end{figure*}

We can see pixel noise, especially from CUT~\cite{park-eccv20}, Pix2PixHD~\cite{wang-cvpr18}, and DoveNet~\cite{cong-cvpr20}, from the last two rows of Fig.~\ref{fig:qualitative_baseline_comparison}.
For CUT and DoveNet, we believe it is because these models do not take the surface normal as their input, thereby losing geometry information in the generation process.
On the other hand, because Pix2PixHD~\cite{wang-cvpr18} takes the normal map as an additional input, it models the human geometry but fails to generate the colors correctly.
It happens because Pix2PixHD models albedo and shading in the same information stream, while our method handles them separately.

Comparably, Deep CG2Real~\cite{bi-iccv19} keeps more details than others, and this is because this approach does not have any of the previously mentioned limitations; it can generate the shaded human while keeping the albedo map intact.
Because this method originally assumes that there is cheap shading input instead of the background image\footnote{We have asked the soundness of this experiment setting to the first author of Deep CG2Real before conducting this comparison and get a positive answer with their currently-private code base.}, however, the network shows difficulties in controlling the brightness, as shown in Fig.~\ref{fig:shading_map_visual_results} and the arms of the first two rows in Fig.~\ref{fig:qualitative_baseline_comparison}.

To better understand~\ourModelName, we further perform an ablation study by removing one loss term in Eq.~\ref{eq:loss_all} and our~\ourRepName~map~$\mathbf{r}^{i}$, and the result is summarized in Table~\ref{tab:quantitative_baseline_comparison}.
First of all, we found that the `no $\mathcal{L}_{s}^{recon}$' model in Table~\ref{tab:quantitative_baseline_comparison}~shows the best performance in MSE. 
We believe the lack of $\ell_{1}$ loss $\mathcal{L}_{s}^{recon}$ leads the model overfits to MSE, based on other losses like the $\ell_{2}$ loss $\mathcal{L}_{g}^{recon}$. 
As a consequence, it performs worse in other metrics like LPIPS. 
Also, given the performance drop of our method when excluding either $\mathbf{r}^{i}$ or $\mathcal{L}^{\text{recon}}_{s}$, which are the major difference between Deep CG2Real~\cite{bi-iccv19}~and our~\ourModelName, $\mathbf{r}^{i}$ and $\mathcal{L}^{\text{recon}}_{s}$ may play a crucial role in obtaining a better shading map.

Lastly, as visualized in each paired example, our model keeps the output temporally coherent, which highlights the robustness of our approach.
We believe our robust self-occlusion representation drives the stability of our method.

\subsection{Parts Segmentation on Real World Images}
\label{subsec:surreal_comparison}

In this section, we show the usefulness of our method beyond harmonization.
We use our trained network to synthesize data to train CNNs for the part segmentation task presented in Varol~\etal~\cite{varol-cvpr17}.
Therefore, we synthesize training data for the segmentation tasks on the Freiburg Sitting People~\cite{oliveira-icra16} and Unite the People~\cite{lassner-cvpr17} datasets, following the same pipeline as in SURREAL~\cite{varol-cvpr17}.

\textbf{Dataset.}
The original SURREAL dataset synthesizes data by employing human motion from CMU MoCap~\cite{cmu-mocap}, body texture from CAESAR~\cite{robinette-afrl02} applied on the SMPL body model~\cite{looper-tog15}, background from LSUN dataset~\cite{yu-arxiv15}, and spherical harmonics with nine coefficients~\cite{green-gdc03}.
We keep the original pipeline as it is (please check the details about dataset rendering from the original SURREAL dataset~\cite{varol-cvpr17} and its project repository).
However, we additionally render the albedo map and foreground alpha over (256, 256) image scale by editing the original rendering pipeline.
We then feed each input to our model by downscaling it to 128 and then upscaling the output back to (256, 256).
The entire dataset synthesized with our method has 67,529 sequences with 100 or fewer frames per sequence identical to the SURREAL CMU set.
We follow the same train, validation, test split as the SURREAL data release (50,806 / 4,195 / 12,528 sequences for train / validation / test, respectively). 
In the rest of the paper, we denote the data synthesized with our method as Synth. 

To evaluate the performance over real-world examples, we tested the CNNs trained with the data synthesized by our method on two datasets.
First, the Freiburg Sitting People (FSitting) dataset~\cite{oliveira-icra16} includes 200 front-view images of (300, 300) from 6 subjects with 14 human parts annotations (upper two rows in Fig.~\ref{fig:part_segmentation_visual_results}).
We follow the same train/test split in the original dataset, which has two subjects for training and four subjects for testing.
Next, we tested the segmentation task from Unite the People (UP) dataset~\cite{lassner-cvpr17}.
Specifically, we used the segmentation dataset, which includes body part segmentation~\cite{shotton-pami12}, on (513, 513), with its own train / validation / test splits (5,703 / 1,423 / 1,389 images).
In our experiments on the UP dataset, we used 3\% of the training data while the validation and test split remained the same.

\textbf{Methodology.}
We train a modified version of the stacked hourglass network architecture employed in SURREAL, initially introduced for 2D pose estimation~\cite{newell-eccv16}.
Specifically, we trained the `upscale' network in SURREAL~\cite{varol-cvpr17} that measures the performance on (256, 256) images for both FSitting and UP.
Thus, we also resize and crop each image into (256, 256) for both training and testing, same as the original work, and train each model over 30 epochs each with a batch size of 4.

We trained the network on (i) synthetic data only, (ii) real data only, (iii) finetuning a model from the case (i) with real data.
We use the mean and standard deviation of each dataset for whitening.
For the synthetic data only cases, we first train on both the original SURREAL dataset (denoted as `SURREAL') and our dataset (`Synth') and then measure the performance on the real dataset independently.
To show the non-finetuned model's performance, along with the finetuned models, in the UP dataset, we equally used 6 part segmentation, as defined in the original UP dataset~\cite{lassner-cvpr17}. 
Please check Appendix~\ref{supp_subsec:up_mapping}~and the GT from the lower two rows in Fig.~\ref{fig:part_segmentation_visual_results} for the details.

For evaluation, we first train three different models, two with the synthetic dataset (case (i)) and one with the real dataset (case (ii)), and finetuned each model trained with the synthetic dataset (case (iii)) using the data from FSitting for 50 iterations, and using the data from UP for 400 iterations.
We report the median performance of those models as the final performance.

\textbf{Results.}
We use intersection over union (IoU) and pixel-level accuracy (PixAcc) as evaluation metrics, following Varol~\etal~\cite{varol-cvpr17}. 
We average the human parts for all settings, and the result is summarized in Table~\ref{tab:part_segmentation_performance}.

First, in the Freiburg Sitting People (FSitting)~\cite{oliveira-icra16} dataset, the performance gap between training with our synthetic data without finetuning (Synth) and training with real data alone is 18.39\%p on PixAcc, but when we train with the SURREAL dataset, the performance gap is 35.38\%p on PixAcc.
We believe this happens because we generate rich shading that is better harmonized with the background, thereby having more realistic synthetic data in comparison to SURREAL.
However, if we do not finetune the model trained with our synthesized data, we observe lower performance than if we train with real data alone.
In Unite the People (UP)~\cite{lassner-cvpr17}, we can see a similar trend on six parts segmentation.

When we move our focus to the finetuned models, we can see that the model pre-trained with our synthesized data shows better performance (16.33\%p higher in FSitting PixAcc, and 7.74\%p higher in UP PixAcc) than training with real data only.
The performance is also higher than the models trained with the data synthesized with SURREAL (5.85\%p gap in FSitting PixAcc, and 1.49\%p gap in UP PixAcc).
We can see better segmentation results in Fig.~\ref{fig:part_segmentation_visual_results} from both datasets as well.
These results show that our synthesized data is closer to the real data distribution in comparison to the data synthesized in SURREAL. 
Data synthesized with our method combined with perfect data labels from the simulation results in improved performance.

\begin{table}[t]
\setlength{\tabcolsep}{5pt}
\centering
\begin{tabular}{l|*2c|*2c}
\toprule
\multirow{2}{*}{Training Data}    & \multicolumn{2}{c|}{FSitting~\cite{oliveira-icra16}} & \multicolumn{2}{c}{UP~\cite{lassner-cvpr17}} \\
                                  & IoU & PixAcc & IoU & PixAcc \\
\midrule
SURREAL                         & 27.07 & 32.51 & 11.02 & 11.79 \\
Synth                           & 38.49 & 49.50 & 21.25 & 25.85 \\\midrule
Real                            & 49.75 & 67.89 & 40.89 & 54.08 \\\midrule
Real + SURREAL                  & 65.99 & 78.37 & 48.45 & 60.33 \\
\textbf{Real + Synth (Ours)}    & \textbf{66.42} & \textbf{84.22} & \textbf{49.55} &\textbf{61.82} \\
\bottomrule
\end{tabular}
\caption{\textbf{Parts segmentation results on Freiburg Sitting People (FSitting)~\cite{oliveira-icra16} and Unite the People (UP)~\cite{lassner-cvpr17}.} The best result is obtained when we finetune the model trained with our synthetic data with real images.} %
\label{tab:part_segmentation_performance}
\end{table}

\section{Conclusion}
\label{sec:conclusion}

We propose~\ourModelName, a deep shading generation framework that harmonizes synthetic humans with a given background image. 
Our main contribution is a novel~\emph{2D self-occlusion map}, named~\ourRepName, that enables understanding shadows caused by body parts occluding each other from the source light without directly feeding any 3D inputs to the network. 
Also, we propose an adversarial learning technique with a set of auxiliary losses to help our network generate better shading maps.
We show our method achieves performance superior to image harmonization and synthetic-to-real approaches, demonstrating the effectiveness of our novel 2D~\ourRepName~map.
More importantly, our method serves as a data generation module that improves the performance of CNNs trained for human body part segmentation. 

Our method does not offer a complete rendering solution at its current state and has limitations we would like to address in future work:
1) Our model does not generate shadows on the background, which is important for generating more realistic images. A potential approach is to jointly perform 3D scene understanding from the background image.
2) The light representation learned by our method is not easily interpretable and thus does not allow for easy control over the outputs of the network. Enabling more control can pave the road to generating data tailored for specific tasks.
3) Our proposed~\ourRepName~\emph{self-occlusion}~map is constructed by a fixed number of rays which may not be sufficient to capture high-frequency details. Exploring continuous representations of the occlusion map is a promising direction.

\clearpage

\appendix
\renewcommand{\thetable}{\Alph{table}}
\renewcommand{\thefigure}{\Alph{figure}}
\renewcommand{\thesection}{\Alph{section}}
\renewcommand{\thesubsection}{\thesection.\alph{subsection}}
\setcounter{figure}{0}
\setcounter{table}{0}

\section*{Appendix}
\label{sec:appendix}

This appendix includes detailed information for the paper, including:
\begin{compactitem}
\item \textbf{Network Architectures} (Section~\ref{supp_sec:architecture}) provide the details about our model architecture $f$, $\mathcal{G}$ and $\mathcal{D}$ (Section~\ref{sec:methods}).
\item \textbf{Dataset} (Section~\ref{supp_sec:dataset}) describes detailed procedures to generate the simulated human motion videos (Section~\ref{subsec:baseline_comparison}).
\item \textbf{Training Details} (Section~\ref{supp_sec:training}) cover the details in training our method (Section~\ref{subsec:baseline_comparison}).
\item \textbf{Additional Experiments} (Section~\ref{supp_sec:additional_experiments}) perform experiments on different models (Section~\ref{subsec:baseline_comparison}). 
\item \textbf{Parts Segmentation Map} (Section~\ref{supp_sec:segmentation_mapping}) explains how the human parts segmentation pairs are matched in each experiment (Section~\ref{subsec:surreal_comparison}).
\end{compactitem}

\section{Network Architectures}
\label{supp_sec:architecture}
\subsection{Light Encoder}
\label{supp_subsec:light_encoder}

For the rest of the tables in this section, `$H \times W \times C$ Conv' means a 2D Convolution layer with $C$ filters size $H \times W$ followed by instance normalization~\cite{ulyanov-cvpr17} and a Leaky ReLU nonlinearity.
We perform a reflection padding before the Conv layer, as introduced in Johnson~\etal~\cite{johnson-eccv16}, not to reduce the activation shape in the table by 1 pixel on each side.
For all upscaling, we use the bilinear upsampling technique introduced in distill.pub\footnote{\url{https://distill.pub/2016/deconv-checkerboard/}}.
If a different activation or normalization is used, then we note such variation in the `Layer' side of a row with a parenthesis.
Detailed architecture of our U-Net~\cite{ronneberger-miccai15} based background to the light representation network (denoted as $f$ in Fig.~\ref{fig:birdeye_view}) is in Table~\ref{supp_tab:bg_to_lightrep_network}.

\begin{table}[t]
\setlength{\tabcolsep}{4pt}
\small
\centering
\begin{tabular}{c|c}
\toprule
Layer & Activation Shape \\
\midrule
  Input  & 128$\times$128$\times$3 \\
  
  5$\times$5$\times$1 Conv, stride 1 (no norm) & 128$\times$128$\times$1 \\
  
  8$\times$8$\times$16 Conv, stride 4 & 32$\times$32$\times$16 \\
  5$\times$5$\times$16 Conv, stride 1 & 32$\times$32$\times$16 \\
  
  8$\times$8$\times$16 Conv, stride 4 & 8$\times$8$\times$16 \\
  5$\times$5$\times$16 Conv, stride 1 & 8$\times$8$\times$16 \\
  
  5$\times$5$\times$16 Conv, stride 1 & 8$\times$8$\times$16 \\
  Bilinear Upsample, scale 4 & 32$\times$32$\times$16 \\
  5$\times$5$\times$16 Conv, stride 1 & 32$\times$32$\times$16 \\
  Channel-wise Concat with row 4 activation & 32$\times$32$\times$32 \\
  
  5$\times$5$\times$8 Conv, stride 1 & 32$\times$32$\times$8 \\
  Bilinear Upsample, scale 4 & 128$\times$128$\times$8 \\
  5$\times$5$\times$8 Conv, stride 1 & 128$\times$128$\times$8 \\
  Channel-wise Concat with row 2 activation & 128$\times$128$\times$9 \\
  
  5$\times$5$\times$16 Conv, stride 1 (no norm, sigmoid) & 128$\times$128$\times$16 \\
\bottomrule
\end{tabular}
\caption{\textbf{Network architecture for $f$.}}
\label{supp_tab:bg_to_lightrep_network}
\vspace{10px}
\begin{tabular}{c|c}
\toprule
Layer & Activation Shape \\
\midrule
  Input  & 128$\times$128$\times$94 \\
  3$\times$3$\times$47 Conv, stride 1 (no norm) & 128$\times$128$\times$47 \\
  
  4$\times$4$\times$8 Conv, stride 2 & 64$\times$64$\times$8 \\
  3$\times$3$\times$8 Conv, stride 1 & 64$\times$64$\times$8 \\
  
  4$\times$4$\times$16 Conv, stride 2 & 32$\times$32$\times$16 \\
  3$\times$3$\times$16 Conv, stride 1 & 32$\times$32$\times$16 \\
  
  4$\times$4$\times$16 Conv, stride 2 & 16$\times$16$\times$16 \\
  3$\times$3$\times$16 Conv, stride 1 & 16$\times$16$\times$16 \\
  
  4$\times$4$\times$16 Conv, stride 2 & 8$\times$8$\times$16 \\
  3$\times$3$\times$16 Conv, stride 1 & 8$\times$8$\times$16 \\
  
  3$\times$3$\times$16 Conv, stride 1 & 8$\times$8$\times$16 \\
  Bilinear Upsample, scale 2          & 16$\times$16$\times$16 \\
  3$\times$3$\times$16 Conv, stride 1 & 16$\times$16$\times$16 \\
  Channel-wise Concat with row 8 activation & 16$\times$16$\times$32 \\
  
  3$\times$3$\times$16 Conv, stride 1 & 16$\times$16$\times$16 \\
  Bilinear Upsample, scale 2          & 32$\times$32$\times$16 \\
  3$\times$3$\times$16 Conv, stride 1 & 32$\times$32$\times$16 \\
  Channel-wise Concat with row 6 activation & 32$\times$32$\times$32 \\
  
  3$\times$3$\times$8 Conv, stride 1 & 32$\times$32$\times$8 \\
  Bilinear Upsample, scale 2         & 64$\times$64$\times$8 \\
  3$\times$3$\times$8 Conv, stride 1 & 64$\times$64$\times$8 \\
  Channel-wise Concat with row 4 activation & 64$\times$64$\times$16 \\
  
  3$\times$3$\times$8 Conv, stride 1 & 64$\times$64$\times$8 \\
  Bilinear Upsample, scale 2         & 128$\times$128$\times$8 \\
  3$\times$3$\times$8 Conv, stride 1 & 128$\times$128$\times$8 \\
  Channel-wise Concat with row 2 activation & 128$\times$128$\times$55 \\
  
  3$\times$3$\times$3 Conv, stride 1 (no norm, sigmoid) & 128$\times$128$\times$3 \\
\bottomrule
\end{tabular}
\caption{\textbf{Network architecture for $\mathcal{G}$.}}
\label{supp_tab:generator_network}
\end{table}

\subsection{Shading Generator}
\label{supp_subsec:shading_generator}

We implement our generator $\mathcal{G}$ based on the BicycleGAN~\cite{zhu-neurips17}, which is based on U-Net~\cite{ronneberger-miccai15}. 
Especially, we take the four Downscale-Upscale pairs for our 128 $\times$ 128 image generation.
Each scaling block has two 2D Convolutions, and we add a smoothing layer before the innermost and after the outermost layer of $\mathcal{G}$.
For the input, we feed the albedo, normal, self-occlusion map, and light representation output from $f$ by channel-wise concatenating them, thereby having $128 \times 128 \times (3+3+62+16) = 128 \times 128 \times 94$ as the input.
For the output, we get a 3 channel shading map $128 \times 128 \times 3$, but with a linear activation, instead of a hyperbolic tangent at the end, to cover the range of shading more precisely. %
Our final generator network $\mathcal{G}$ is in Table~\ref{supp_tab:generator_network}.

\begin{table}[t]
\small
\centering
\begin{tabular}{c|c}
\toprule
Layer & Activation Shape \\
\midrule
  Input  & 128$\times$128$\times$6 \\
  3$\times$3$\times$3 Conv, stride 1 & 128$\times$128$\times$3 \\
  4$\times$4$\times$8 Conv, stride 2 & 64$\times$64$\times$8 \\
  4$\times$4$\times$16 Conv, stride 2 & 32$\times$32$\times$16 \\
  4$\times$4$\times$16 Conv, stride 2 & 16$\times$16$\times$16 \\
  4$\times$4$\times$16 Conv, stride 2 & 8$\times$8$\times$16 \\
\bottomrule
\end{tabular}
\caption{\textbf{Network architecture for $\phi$ in $\mathcal{D}$.}}
\label{supp_tab:discriminator_phi_network}
\vspace{10px}
\begin{tabular}{c|c}
\toprule
Layer & Activation Shape \\
\midrule
  Input  & 8$\times$8$\times$16 \\
  1$\times$1$\times$8 Conv, stride 1 & 8$\times$8$\times$8 \\
  1$\times$1$\times$1 Conv, stride 1 (no norm, linear) & 8$\times$8$\times$1 \\
\bottomrule
\end{tabular}
\caption{\textbf{Network architecture for $\psi$ in $\mathcal{D}$.}}
\label{supp_tab:discriminator_psi_network}
\vspace{10px}
\begin{tabular}{c|c}
\toprule
Layer & Activation Shape \\
\midrule
  Input  & 128$\times$128$\times$16 \\
  5$\times$5$\times$8 Conv, stride 2 & 128$\times$128$\times$8 \\
  8$\times$8$\times$16 Conv, stride 4 & 32$\times$32$\times$16 \\
  4$\times$4$\times$16 Conv, stride 4 (no norm, linear) & 8$\times$8$\times$16 \\
\bottomrule
\end{tabular}
\caption{\textbf{Network architecture for $\omega$ in $\mathcal{D}$.}}
\label{supp_tab:discriminator_omega_network}
\end{table}

\subsection{Shading Discriminator}
\label{supp_subsec:shading_discriminator}

Our conditional discriminator ($\mathcal{D}$ in Fig.~\ref{fig:birdeye_view}) is based on the projection discriminator introduced in Miyato~\etal~\cite{miyato-iclr18}. 
Therefore, our discriminator network consists of a main shared stream network $\phi$ ~(Table~\ref{supp_tab:discriminator_phi_network}), a main downstream network $\psi$~(Table~\ref{supp_tab:discriminator_psi_network}), and a pre-projection transformation network $\omega$~(Table~\ref{supp_tab:discriminator_omega_network}). 
We described each of them in Tables~\ref{supp_tab:discriminator_phi_network},~\ref{supp_tab:discriminator_psi_network}, and~\ref{supp_tab:discriminator_omega_network}. 
The only difference in policy is that we replace the default normalization method used in $\mathcal{G}$ from instance normalization~\cite{ulyanov-cvpr17} to spectral normalization~\cite{miyato-iclr18-spectral} for a more stable GAN~\cite{goodfellow-neurips14} training.

Once we have all those three subnetworks $\phi$, $\psi$, and $\omega$, we can generate output by $\psi(\phi(\mathbf{y}::\mathbf{o})) + \phi(\mathbf{y}::\mathbf{o}) \cdot \omega(\mathbf{l})$, where $\left[::\right]$ means channel-wise concatenation and $\left[\cdot\right]$ means channel-wise dot-product.
Because the shading can theoretically go to infinity, we pass a hyperbolic tangent before feeding it to $\phi$ to prevent overflow, thus $\mathbf{y} = \text{tanh}(\mathbf{y}_{s})$~or~$\text{tanh}(\mathbf{\hat{y}}_{s})$.
We empirically found that adopting the GAN-CLS technique introduced by Reed~\etal~\cite{reed-icml16} gives a more precise signal to each conditional input of $\mathcal{D}$, so we employ this term in the generator gradient update phase throughout the experiment.
We would like to note that we take the PatchGAN~\cite{isola-cvpr17} approach, however, we provide different conditions per each region to guide the fine-grain details in shading.

\section{Dataset}
\label{supp_sec:dataset}

We construct a large-scale dataset by retargeting 398 randomly selected motion sequences from~Mixamo~\cite{adobe-mixamo} into 38 3D human models in RocketBox~\cite{gonzalez-fvr20}.
The materials of the human body in our dataset are composed of Oren-Nayar diffuse Bidirectional Scattering Distribution Function (BSDF)~\cite{oren-siggraph94} and microfacet glossy BSDF~\cite{walter-eg07}.
We randomly pick one of 8 IBL textures downloaded from HDRI Haven~\cite{hdri-haven}.
We also place an infinite virtual ground plane under the foreground human body to catch the shadows underneath with a differential rendering algorithm~\cite{debevec-siggraph98}.
Then, we put three evenly distributed cameras in front of the human body for each animated sequence.
Each camera is set to track the character's chest in the center, and its trajectory is further smoothed to avoid abrupt camera motion. 
The camera distance is also slightly perturbed to bring more diversity. 
The final images are rendered with Monte Carlo path tracing~\cite{kajiya-siggraph86}.
As a result, we obtain a total of 212,955 frames.

\section{Training Details}
\label{supp_sec:training}
To train our network, we first build our pipeline after preprocessing the self-occlusion maps.
Specifically, for faster ray-marching in obtaining a self-occlusion map, we employ the `pyembree' library.
Then, in the data pipeline construction stage, we slice each rendered motion sequence by 10 frames from three different camera positions for training and 30 frames from a single camera position for testing.
We slide the slicing window by half-sequence (5 frames for training and 15 frames for testing) and drop the sequence with fewer frames.
We directly load the precomputed self-occlusion maps for a faster data serving because it will not change once the camera position and the object pose are fixed.

For a more stable training, we scaled the edge gradient mask $\mathbf{x}_{w}$ by dividing the maximum value per channel, thereby keeping the value to be in a predictable range.
Also, for the feature matching loss in Eq.~3, we compute the feature distance from the second and third Conv output from the $\phi$ network only, similar to Jang~\etal~\cite{jang-icml18}, because we practically see that it disturbs the training once we add the loss from deeper layers.
Lastly, we set 10 for $\mathcal{L}^{\text{recon}}_{g}$, 1,000 for $\mathcal{L}^{\text{FM}}_{s}$, and all the rest to 1 in training the network.

\section{Additional Experiments}
\label{supp_sec:additional_experiments}
\begin{table}[t]
\setlength{\tabcolsep}{2pt}
\small
\centering
\begin{tabular}{l|*5c}
\toprule
\multirow{2}{*}{Method}       & MSE$\downarrow$ & LPIPS$\downarrow$ & SSIM $\uparrow$ & FVD$\downarrow$ & MOVIE$\downarrow$ \\
                              & {\scriptsize($\times 10^{-3}$)} & {\scriptsize($\times 10^{-2}$)} & {\scriptsize($\times 10^{-1}$)} & {\scriptsize($\times 10^{1}$)} & {\scriptsize($\times 10^{-4}$)} \\
\midrule
CUT~\cite{park-eccv20}        & 11.68 & 2.93 & 9.68 & 9.99 & 22.39 \\
Pix2PixHD~\cite{wang-cvpr18}  &  9.89 & 3.01 & 9.65 & 9.10 & 17.10 \\
DoveNet~\cite{cong-cvpr20}    &  7.72 & 2.13 & 9.72 & 12.47 & 19.48 \\
RainNet~\cite{ling-cvpr21}    &  8.88 & 2.40 & 9.70 & 13.59 & 22.08 \\
BargainNet~\cite{cong-icme21} &  10.10 & 2.78 & 9.65 & 14.53 & 21.30 \\
Deep CG2Real~\cite{bi-iccv19} &  6.50 & 2.19 & 9.71 & 10.11 & 11.10 \\
\textbf{\ourModelName~(full)} & \textbf{3.72}  & \textbf{1.24} & \textbf{9.87} & \textbf{4.61} & \textbf{7.00} \\
\bottomrule
\end{tabular}
\vspace{6pt}
\caption{\textbf{Quantitative result with RainNet~\cite{ling-cvpr21} and BargainNet~\cite{cong-icme21}.} We additionally measure the visual quality of RainNet~\cite{ling-cvpr21} and BargainNet~\cite{cong-icme21}, following the same protocol as in Section~\ref{subsec:baseline_comparison}. 
$\uparrow$ next to the evaluation metric means a higher number is favored for that metric, and $\downarrow$ is for the opposite, same as Table~\ref{tab:quantitative_baseline_comparison}} 
\label{supp_tab:quantitative_comparison}
\end{table}

We additionally measure the performance in RainNet~\cite{ling-cvpr21} and BargainNet~\cite{cong-icme21}, following the same protocol as in Section~\ref{subsec:baseline_comparison}. 
The results of those models are summarized in Table~\ref{supp_tab:quantitative_comparison}.
First of all, those models show worse performance than ours.
It seems that the nonexistence of the surface normal information causes models to produce pixel noise, similar to CUT~\cite{park-eccv20} and DoveNet~\cite{cong-cvpr20}.

Surprisingly, those two models show even weak performance than DoveNet~\cite{cong-cvpr20}, which those works set as one of the baseline approaches.
We believe our background environment may be inadequate to be represented as a single vector for either normalization~\cite{ling-cvpr21} or as a context vector~\cite{cong-icme21}.

Inspired by those results, we design~\ourModelName~to feed our representation context in a pixel-aligned form, which comes from a convolution network $f$, as in Section~\ref{supp_sec:architecture}.
Thereby, the nearby input can still get feedback via the convolution operation while focusing more on individual pixels.

\section{Parts Segmentation Map}
\label{supp_sec:segmentation_mapping}
In the human parts segmentation task, we use various datasets: SURREAL~\cite{varol-cvpr17} (both original SURREAL dataset, denoted as SURREAL, and an additional dataset with our shading, but with the same 3D human, motion, and background from SURREAL, named as Synth), Freiburg Sitting (FSitting)~\cite{oliveira-icra16}, and Unite the People (UP)~\cite{lassner-cvpr17}.
The original SURREAL and Synth dataset includes the same number of data.
Because each dataset segmented a human into a different number of parts (24 for SURREAL / 14 for FSitting / 31 for UP), we set each pair in the experiment to have the same number of labels to measure non-finetuned models' performance.
Although the final parts are visualized in Fig.~\ref{fig:part_segmentation_visual_results}, we provide a matching table for each pair for better reproducibility below.

\subsection{Evaluation on Freiburg Sitting Dataset}
\label{supp_subsec:fsitting_mapping}

\begin{table}[t]
\setlength{\tabcolsep}{4pt}
\small
\centering
\begin{tabular}{rl|c|c}
\toprule
Index & Parts Name & SURREAL & FSitting \\
\midrule
  1 & Head            &                      16 &   1 \\
  2 & Torso           & 1, 4, 7, 10, 13, 14, 15 &   2 \\
  3 & RightUpperArm   &                      18 &   4 \\
  4 & RightLowerArm   &                      20 &   3 \\
  5 & RightHand       &                  22, 24 &   5 \\
  6 & LeftUpperArm    &                      17 &   7 \\
  7 & LeftLowerArm    &                      19 &   6 \\
  8 & LeftHand        &                  21, 23 &   8 \\
  9 & RightUpperLeg   &                       3 &  10 \\
 10 & RightLowerLeg   &                       6 &   9 \\
 11 & RightFoot       &                   9, 12 &  11 \\
 12 & LeftUpperLeg    &                       2 &  13 \\
 13 & LeftLowerLeg    &                       5 &  12 \\
 14 & LeftFoot        &                   8, 11 &  14 \\
\bottomrule
\end{tabular}
\caption{\textbf{Mapping table between SURREAL~\cite{varol-cvpr17} and FSitting~\cite{oliveira-icra16} dataset.} The segmentation indexes in each dataset (columns 3 and 4) are converted to the first column index.}
\label{subtab:surreal_fsitting_mapping}
\end{table}

\begin{table}[t]
\setlength{\tabcolsep}{2pt}
\small
\centering
\begin{tabular}{rl|c|c}
\toprule
Index & Parts Name & SURREAL & UP \\
\midrule
  1 & Head           &                      16 &   27, 28, 29, 30 \\
  2 & Torso          & 1, 4, 7, 10, 13, 14, 15 &   6, 7, 8, 19, 20, 21, 31 \\
  3 & RightArm       &          18, 20, 22, 24 &   14, 15, 16, 17, 18 \\
  4 & LeftArm        &          17, 19, 21, 23 &   1, 2, 3, 4, 5 \\
  5 & RightLeg       &             3, 6, 9, 12 &   22, 23, 24, 25, 26 \\
  6 & LeftLeg        &             2, 5, 8, 11 &   9, 10, 11, 12, 13 \\
\bottomrule
\end{tabular}
\caption{\textbf{Mapping table between SURREAL~\cite{varol-cvpr17} and UP~\cite{lassner-cvpr17} dataset.} The segmentation indexes in each dataset (columns 3 and 4) are converted to the first column index.}
\label{subtab:surreal_up_mapping}
\end{table}

Because we can convert the SURREAL~\cite{varol-cvpr17} to Freiburg Sitting (FSitting)~\cite{oliveira-icra16} dataset by directly merging few parts in SURREAL, we use the human parts segmentation predefined in FSitting as the target. 
Therefore, we have a total of 14 parts for both datasets.
Table~\ref{subtab:surreal_fsitting_mapping} shows how we map each part segmentation label in SURREAL and FSitting to have the same index.
Since we use the predefined 14 segmentation mask mapping in the original SURREAL project repository\footnote{\url{https://github.com/gulvarol/surreal}}, we flipped the arm and leg index in FSitting as well, as in Table~\ref{subtab:surreal_fsitting_mapping}.

\subsection{Evaluation on Unite the People Dataset}
\label{supp_subsec:up_mapping}

Similar to the Freiburg Sitting Dataset case, the SURREAL~\cite{varol-cvpr17} dataset's segmentation parts are not matched with the Unite the People (UP)~\cite{lassner-cvpr17} Dataset.
Thus we make both datasets have the same number of segmentation parts.
Specifically, because the human parts segmentation of the UP dataset cannot be matched with the FSitting's 14 segmentation map, we instead merge the segmented parts from two datasets into 6 parts, as predefined in the UP dataset.
Table~\ref{subtab:surreal_up_mapping} shows the final 6 human segmentation parts mapping from each dataset.

\medskip
\textbf{Acknowledgements.} %
This work was supported in part by Kwanjeong Educational Foundation Scholarship. We thank Junhyug Noh, Chris Dongjoo Kim, Jongwook Choi and Lajanugen Logeswaran for constructive feedback of the manuscript.

\clearpage
{\small
\bibliographystyle{ieee_fullname}
\bibliography{aicc2022}
}

\end{document}